%% file: main.tex
\documentclass{article}
\usepackage{iclr2027_conference,times}

\input{math_commands.tex}

\usepackage{hyperref}
\hypersetup{colorlinks=true,linkcolor=blue!60!black,citecolor=blue!60!black,urlcolor=blue!60!black,pdftitle={Priming Base Models for Agentic Post-Training: An Open Midtraining Recipe},pdfauthor={NVIDIA}}
\usepackage{url}
\usepackage{graphicx}
\usepackage{booktabs}
\usepackage{multirow}
\usepackage{xcolor}
\usepackage{colortbl}
\usepackage{wrapfig}
\definecolor{gaingreen}{RGB}{0,125,60}
\newcommand{\gain}[1]{\textcolor{gaingreen}{+#1}}
\usepackage{amsmath,amssymb}
\usepackage{tikz}

\newcommand{\nanofull}{Nemotron~3.5 Lightning}
\newcommand{\nano}{N-3.5 Lightning}
\newcommand{\nanolc}{N-3.5 Lightning-LC}
\newcommand{\nanoeightk}{N-3.5 Lightning-8K}
\newcommand{\superfull}{Nemotron~3.5 Super}
\newcommand{\superm}{N-3.5 Super}
\newcommand{\superintro}{the larger \superfull{}-120B-A12B}
\newif\ifarxiv\arxivtrue

\title{Priming Base Models for Agentic\\ Post-Training: An Open Midtraining Recipe}

\author{\small Soumye Singhal$^{*}$, Joseph Jennings$^{*}$, Ka Shun Shum$^{*\dagger}$, Alexander Bukharin$^{*}$, Khushi Bhardwaj$^{*}$\\
\small\textbf{Zirui Liu, Venkat Srinivasan, Jonathan Li, Yash Jain, Kezhi Kong, Markus Kliegl,}\\
\small\textbf{Olivier Delalleau, Sanjeev Satheesh, Mostofa Patwary}\\[4pt]
\normalfont NVIDIA}

\iclrfinalcopy
\makeatletter
\newcommand{\firstpagenote}[1]{{\renewcommand\@makefntext[1]{\noindent##1}\renewcommand{\thefootnote}{}\footnotetext{#1}}}
\makeatother
\begin{document}

{\setlength{\tabcolsep}{1pt}\maketitle}
\firstpagenote{$^{*}$Core contributors. Correspondence: \texttt{\{jjennings,\,soumyes,\,abukharin\}@nvidia.com}}
\firstpagenote{$^{\dagger}$Work done during an internship at NVIDIA.}
\lhead{Preprint}

\begin{abstract}
Frontier models, general-purpose and domain-specific alike, now include a midtraining stage between pretraining and post-training, but little has been published about how this stage is built. We present an open midtraining recipe that primes base models for agentic post-training, and we show its effectiveness on two Nemotron models of different sizes. Pretraining typically runs at short context and sees almost no agent interactions, while SFT data is mostly synthetic, and very long SFT stages can hurt the model's later RL trainability. Our midtraining stage sits between the two. It trains at 256K context for hundreds of billions of tokens on a diverse mixture of natural software-engineering data like packed repositories, pull requests, commit histories, and build logs together with a small share of agentic traces. On \nanofull{}, 500B midtraining tokens improve SWE-bench Verified from 60.7 to 64.5, SWE-bench Multilingual from 55.8 to 62.8, and SWE-bench Pro from 32.0 to 35.5 after identical full SFT. On the larger \superfull{}, 280B midtraining tokens improve DeepSWE, a harder benchmark of long-horizon tasks, from 11.7 to 15.7 and SWE-bench Verified from 72.9 to 74.3 after identical full SFT. Midtraining also improves robustness across agent harnesses on both models. We report the ablations behind the recipe and an evaluation protocol built on short fixed SFT runs, since base-model proxy metrics did not correlate reliably with post-training results.

\end{abstract}

\input{sections/1_introduction}
\input{sections/2_related_work}

\input{sections/3_recipe}
\input{sections/4_results}
\input{sections/5_analysis}

\input{sections/6_discussion}

\subsubsection*{Acknowledgments}
We thank Perplexity, a partner in the Nemotron Coalition, for contributing agentic trace data to the midtraining blend, and in particular Dongqi Wu, Chuan Wang, Bo Wang, and Hang Wu. We also thank the NVIDIA Nemotron pretraining and post-training teams, whose models, data, and training stack this work builds on. We especially thank Luis Vega for the agent-harness sweeps, Krishna Puvvada and Abhinav Khattar for help with the long-context and pretraining data setup, Wasi Uddin Ahmad and Sugam Devare for agentic and software-engineering data, and Jiantao Jiao and Zhenghao Xu for discussions on base-model evaluation. We are grateful to Igor Gitman, Somshubra Majumdar, Makesh Narsimhan Sreedhar, Surya Nistala, Ritu Gala, Abhibha Gupta, and Tan Yu for valuable discussions and help. We thank Bryan Catanzaro, Jonathan Cohen, Oleksii Kuchaiev, and Mohammad Shoeybi for their support, and the compute and infrastructure teams who made this work possible.

\bibliography{references}
\bibliographystyle{iclr2027_conference}

\appendix
\input{sections/7_appendix}

\end{document}

%% file: math_commands.tex
\usepackage{amsmath,amsfonts,bm}

\def\eqref#1{equation~\ref{#1}}

\def\1{\bm{1}}

\DeclareMathAlphabet{\mathsfit}{\encodingdefault}{\sfdefault}{m}{sl}
\SetMathAlphabet{\mathsfit}{bold}{\encodingdefault}{\sfdefault}{bx}{n}



%% file: sections/1_introduction.tex
\section{Introduction}
\label{sec:intro}

Between pretraining and post-training, frontier coding models such as Qwen3-Coder-Next \citep{qwen3codernext}, GLM-5 \citep{glm5}, Kimi-Dev \citep{kimidev}, and Composer~2 \citep{composer2} include a \emph{midtraining} stage of hundreds of billions of tokens of agentic, long-context data on top of a converged base model. Technical reports describe this stage at a high level, but rarely explain how its data is constructed, filtered, and weighted, or measure the impact of each design choice. Following Qwen-Coder \citep{hui2024qwen25coder,qwen3codernext}, we present an open midtraining recipe for agentic coding, developed and demonstrated on Nemotron base models. We describe how each data source is built, and we report the ablations that guided our training choices.

Base models are usually pretrained at short sequence lengths, typically 4K to 8K tokens for compute efficiency \citep{deepseekv3,qwen3}, and then extended to long context in a short final stage before SFT. Nemotron models, for example, are pretrained at 8K and extended to 1M. The data that matters most for software-engineering agents does not fit this pipeline. Pretraining corpora usually contain almost no agent interactions, as in the Nemotron pretraining corpus \citep{nemotron3}, and the natural data around software engineering, such as whole repositories, code evolution through commit histories, pull requests, and build logs, often runs to tens or hundreds of thousands of tokens, too long for pretraining and too large for a short long-context stage. Adding agent traces to pretraining is also not straightforward, since pretraining typically applies no loss masking, and it is unclear how the tool responses (environment outputs) in those traces should be handled. SFT in turn consists mostly of synthetic chat data. Recent work has documented that very long SFT stages can reduce how much the model later gains from RL \citep{rejuvenation,quagmires,goodsft}, and we have observed this in our own training as well.

Midtraining fills this gap. Before the final long-context stage, it trains for hundreds of billions of tokens at 256K, long enough to hold whole repositories, commit histories, and long agent traces (\S\ref{sec:lc}). The data is mostly natural software-engineering data, with a bounded share of agent traces whose tool responses are masked from the loss. Much of this learning can therefore move out of SFT, so SFT can stay short and preserve the model's later RL trainability \citep{rejuvenation,quagmires,goodsft}.

\ifarxiv We develop the recipe on \nanofull{}-30B-A3B and then scale it to \superintro{} \citep{nemotron35}.\else We develop the recipe on \nanofull{}-30B-A3B \citep{nemotronlightning} and then scale it to \superintro{}.\fi{} On \nano{}, 500B midtraining tokens improve SWE-bench Verified, Multilingual, and Pro by 3.8, 7.0, and 3.5 points over the same model without midtraining after identical full SFT, with the most sustained gains on SWE-bench Multilingual (\S\ref{sec:scaling}). On \superm{}, 280B midtraining tokens improve DeepSWE from 11.7 to 15.7 and SWE-bench Verified from 72.9 to 74.3 after identical full SFT (\S\ref{sec:super}). Midtraining also improves robustness to the agent harness, which matters for open models that are used with many different harnesses. Both models improve on every harness we tested, and on \superm{} after 65B tokens and short SFT, the average gain across harnesses is 8.6 points on SWE-bench Verified and 6.8 on Terminal-Bench (\S\ref{sec:harness}).

We make the following contributions.
\begin{itemize}\setlength{\itemsep}{0pt}\setlength{\topsep}{2pt}
  \item A fully documented midtraining data recipe, showing that a small share of agentic traces adds clear gains on top of natural data (\S\ref{sec:recipe} and \S\ref{sec:nano}).
  \item An evaluation protocol based on identical SFT runs, since base-model proxy metrics such as bits-per-byte and LoCoDiff did not reliably predict post-training results (\S\ref{sec:protocol}).
  \item We show that gains hold as we scale both model size and midtraining tokens up to 500B, across benchmarks and agent harnesses (\S\ref{sec:results}).
  \item Ablations of modeling choices, such as the learning rate, the share of replayed pretraining data, tool-response masking, and midtraining's effect on long-context ability (\S\ref{sec:analysis}).
\end{itemize}

%% file: sections/2_related_work.tex
\section{Related Work}
\label{sec:related}

\paragraph{Midtraining in frontier coding models.}
Although midtraining is growing in popularity, only a few works have presented detailed recipes and ablation results to the community. Qwen3-Coder-Next provides one such recipe, midtraining at a 256K context on trillions of tokens, including 600B tokens of repository-level data across 370 programming languages \citep{qwen3codernext}. Its recipe suggests combining mined PR data with agentic traces but does not explain exactly how to do so. GLM-5 reports midtraining on approximately 1.55T tokens with progressive context extension, including 160B tokens of software-engineering data constructed from roughly 10M issue–PR pairs \citep{glm5}, but does not provide details of the recipe. Kimi-Dev, one of the primary works on which we build, provides a more detailed recipe \citep{kimidev}. It performs midtraining on approximately 150B tokens of diff patches, commit sequences, and synthetic agent interactions, and shows that a model with this prior needs far fewer SFT tokens than the raw base model to reach the same agent performance. daVinci-Dev advocates trajectory-style midtraining data and matches Kimi-Dev’s performance using fewer than half as many tokens \citep{davincidev}. Composer~2 shows that code-focused continued pretraining improves downstream RLVR performance \citep{composer2}. AgentFounder similarly shows that continued pretraining benefits search agents \citep{agentfounder}, while \citet{midtrainingbridges} study when midtraining effectively bridges the distributions of pretraining and post-training data. Relative to these works, we provide the ablation-level detail missing from the technical reports and evidence that the gains persist through full SFT.

\paragraph{Repository-level data construction.}
StarCoder2 concatenates repository files in random order \citep{lozhkov2024starcoder2}; DeepSeek-Coder orders files by import dependencies \citep{guo2024deepseekcoder}; SPLiCe shows that structured packing improves long-context use \citep{splice}; ProLong randomizes on purpose to avoid recency bias \citep{prolong}. \citet{jetbrains2025repo} find that most of the gain from repository-level pretraining for project-level code completion comes from adapting to a longer context rather than from how the repository context is assembled. Best-Fit Packing reduces truncation harms \citep{ding2024fewer}, and In-Context Pretraining shows related-document packing helps beyond code \citep{shi2024incontext}. Our repository data differs in \emph{selection}. We keep only repositories whose files reference each other through imports, and order files so that related code is adjacent (\S\ref{sec:recipe}).

\paragraph{Evaluating base models for agentic potential.}
Agentic benchmarks are unreliable before post-training, so Composer~2 selects among base models using held-out codebase perplexity, state tracking, and adversarially built QA instead \citep{composer2}. Other benchmarks target specific skills. LoCoDiff measures file-state reconstruction from git history \citep{locodiff}, RepoBench and CrossCodeEval measure cross-file completion \citep{repobench,crosscodeeval}, and APTBench converts agent tasks into a format base models can answer \citep{aptbench}. Whether such base-model measures predict downstream ability is still contested \citep{samloss,flpm}. In our experiments, they did not track post-training results reliably (\S\ref{sec:protocol}).

%% file: sections/3_recipe.tex
\section{The Midtraining Recipe}
\label{sec:recipe}

\begin{figure}[t]
\centering
\begin{tikzpicture}[xscale=0.122, yscale=1.0]
  \fill[blue!55]   (0,0)  rectangle (75,0.8);
  \fill[orange!70] (75,0) rectangle (95,0.8);
  \fill[teal!60]   (95,0) rectangle (100,0.8);
  \node[white] at (37.5,0.4) {\small Midtraining data \ 75\%};
  \node[white] at (85,0.4)   {\small Replay \ 20\%};
  \draw[gray] (97.5,0.8) -- (97.5,0.95);
  \node[anchor=south east,inner sep=1pt] at (100,0.95) {\scriptsize Long context 5\%};
  \node[anchor=east] at (-1,0.4) {\small Blend};
  \fill[blue!70] (0,-1.2)     rectangle (54.37,-0.4);
  \fill[blue!50] (54.37,-1.2) rectangle (81.04,-0.4);
  \fill[blue!35] (81.04,-1.2) rectangle (94.37,-0.4);
  \fill[blue!20] (94.37,-1.2) rectangle (98.38,-0.4);
  \fill[blue!10] (98.38,-1.2) rectangle (100,-0.4);
  \node[white] at (27.2,-0.8)  {\small Issue--PR 41\%};
  \node[white] at (67.7,-0.8)  {\small Repos 20\%};
  \node[white] at (87.7,-0.8)  {\scriptsize Traces 10\%};
  \draw[gray] (96.4,-1.2) -- (88.5,-1.35);
  \draw[gray] (99.2,-1.2) -- (99.2,-1.35);
  \node[anchor=north east,inner sep=1pt] at (88.5,-1.35) {\scriptsize Diff 3\%};
  \node[anchor=north east,inner sep=1pt] at (100,-1.35) {\scriptsize Logs 1\%};
  \node[anchor=east] at (-1,-0.8) {\small Sources};
  \draw[dashed,gray] (0,-0.4) -- (0,0);
  \draw[dashed,gray] (100,-0.4) -- (75,0);
\end{tikzpicture}
\caption{The midtraining blend, by share of training tokens. Replay is drawn from the base model's pretraining data. Token counts and epochs per source are in \ifarxiv Table~\ref{tab:blend}\else Appendix~\ref{app:blend}\fi.}
\label{fig:blend}
\vspace{-3mm}
\end{figure}
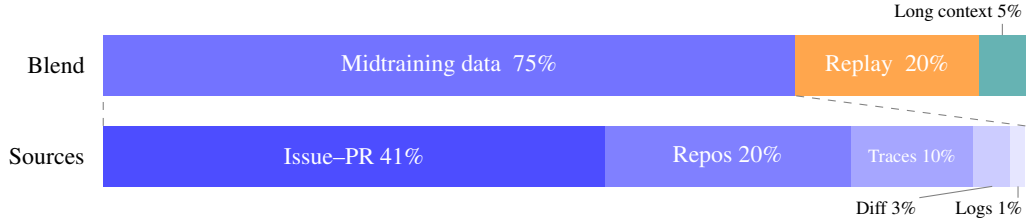

\ifarxiv
\input{sections/blend_table}
\fi

Figure~\ref{fig:blend} summarizes the blend, which is designed for diversity. Most of it is natural software-engineering data, with a small share of synthetic agent traces. We render these traces in the Nemotron chat template and keep them to 10--15\% of the blend to avoid overfitting to SFT-like data, which can hurt later RL trainability \citep{rejuvenation,quagmires,goodsft}. Everything else stays in natural formats, without chat templates. Every source keeps only permissive licenses and goes through decontamination against SWE-bench and Terminal-Bench, secret redaction, and loss masking of content the model should not imitate, such as repeated delimiters, hashes, and timestamps.

\paragraph{Issue--PR.}
We construct a large-scale code-change corpus from merged pull requests in public GitHub repositories, combining repository snapshots with PR descriptions, linked issues, review discussions, and observed patch evolution when available. We restrict collection to repositories with at least ten stars in supported programming languages. We deduplicate examples by repository and PR identifier and remove records with unavailable repositories, missing Git objects, failed patch reconstruction, empty or excessively large changes, generated content, or other low-quality patches. Each example is laid out as a sequence in which repository metadata, date, file tree, and code context condition the model’s reconstruction of the submitted changes. No chat template is applied to this data, but we mask repetitive formatting text from the training loss. In total, this process leaves us with 12.1 million PRs and 200 billion to 2 trillion tokens depending on how much repository context we include in each example. For issue-linked PRs, a structured LLM judge additionally checks that the issue and the patch match, and PR-only examples are filtered with deterministic rules (Appendix~\ref{app:issuepr}).

\paragraph{Repositories.}
Pretraining corpora have historically presented code as individual files rather than whole repositories, and even when files from one repository are packed together, a short pretraining context holds only a small part of it. Agents, however, work inside repositories far longer than that and must use code defined far from where it is called. Midtraining is where the model first reads whole repositories and predicts code that depends on files tens of thousands of tokens earlier. This helps most when the files actually depend on each other, and in many repositories they do not, since many are collections of scripts, exercises, or snippets that never import one another. We start from the repositories in the Nemotron 3 pretraining corpus \citep{nemotron3}, build an import graph for every repository in 14 languages, and keep only repositories in which at least 20\% of code files have an import edge to another file in the repository (10\% for Ruby). Kept repositories are serialized with the top-level README first and connected files next to each other, and packed at 256K, with longer repositories split between connected components rather than truncated, giving about 104B tokens. Compared to alphabetical ordering, dependency ordering better separates stronger from weaker models. It lowers bits-per-byte for stronger models and raises it for weaker ones, which suggests that it exposes cross-file structure that weaker models have not yet learned. The languages, filters, and blend weights are listed in Appendix~\ref{app:repos}.

\paragraph{Diff data.}
Much of software engineering is changing existing code, such as making an edit that stays consistent across several files or keeping track of what a file looks like after a long series of changes. Repository snapshots only show code in its final state, so we build two kinds of examples from commit histories. \emph{Multi-file edits} take a merged pull request, show the model the affected files before the change, and train it to produce them after the change. We keep only changes whose files are linked in the import graph used for the repository data, so that the edits are coupled. \emph{Code-evolution documents} give the initial state of a file, the sequence of real commits that changed it from oldest to newest, and its final state. This trains the long-range state tracking that LoCoDiff measures \citep{locodiff}. Histories range from a single commit to several dozen. Diff headers, commit hashes, authors, and dates are masked from the loss. Sources, filters, and statistics are in Appendix~\ref{app:diff}.

\paragraph{Build and CI logs.}
Agents constantly read logs, stack traces, and build output, but PR data lacks the environment's response, so we add raw system logs from the full LogHub archives \citep{loghub} and CI/CD logs from GitHub Actions run logs and workflow-file histories \citep{ghalogs,workflowhistories}, plus a small curated slice pairing log windows with diagnoses (Appendix~\ref{app:logs}).

\paragraph{Agentic traces.}

We include agent trajectories to teach interaction patterns that natural
software artifacts lack, such as multi-step tool use, testing changes, and
recovering from failures. These trajectories are separate from the SFT data.
The highest-quality trajectories are reserved for SFT, and midtraining puts
the rest to use, including trajectories from earlier teacher models and
tasks that do not fit post-training.

The corpus used in our blend contains about 39B tokens (over one million trajectories, about 500K distinct queries, and 30M tool calls, including SWE-Zero and SWE-Hero trajectories \citep{swezero}).\ifarxiv{} A small portion of the agentic midtraining blend consists of data contributed by Perplexity as a partner in the Nemotron Coalition.\fi{} The full trace corpus covers more than ten agent scaffolds, including OpenHands \citep{openhands}, OpenCode, SWE-agent \citep{sweagent}, Codex, Claude Code, and Terminus, across software engineering, terminal use, search, and general tool use, because traces from a single scaffold transfer poorly to others \citep{qwen3codernext}. Each trajectory keeps its scaffold's instructions and tool definitions and is rendered in the Nemotron chat template. Appendix~\ref{app:traces} gives the statistics and distributions.

Filtering targets shortcuts and undesirable interaction patterns,
including solution leakage through Git history or Internet lookup,
unfinished trajectories, malformed tool calls, and repetitive
file-inspection loops. As in SFT, we train only on assistant turns,
including tool calls, since also training on tool responses hurts results
after SFT (\S\ref{sec:masking}).

\subsection{Training}
\label{sec:training}

Midtraining starts from a converged base model, so we choose the training setup to add new capabilities without degrading what the base model already knows. We train at 256K context, long enough to hold almost every midtraining document, many of which are far longer than the 8K pretraining context (Figure~\ref{fig:doc-lengths}). Our first \nano{} experiments start from the 8K pretraining checkpoint, \nanoeightk{}. For the \nano{} scaling run and all \superm{} runs, we start from the 1M long-context checkpoint, which lets us go straight from midtraining to SFT, keeping evaluations simple. In a mature pipeline, midtraining would come before the final long-context stage (\S\ref{sec:lc}). We set the replay share to 20\% (\S\ref{sec:replay}). The learning rate warms up briefly and then stays constant, at 5e-5 for \nano{}, which favors agentic gains, and 1e-5 for the larger \superm{}, which favors preserving general capabilities (\S\ref{sec:lr}). Appendix~\ref{app:training} lists the full settings.

%% file: sections/blend_table.tex
\begin{table}[t]
\centering
\small
\begin{tabular}{lrrrr}
\toprule
Source & Unique tokens & Share of blend & Tokens seen (500B run) & Epochs \\
\midrule
Issue--PR          & 216.9B & 40.8\% & 205.2B & 0.95 \\
Repositories       & 104.1B & 20.0\% & 100.7B & 0.97 \\
Agentic traces     &  39.4B & 10.0\% &  50.3B & 1.28 \\
Diff data          &  16.0B &  3.0\% &  15.1B & 0.95 \\
Build and CI logs  &   6.5B &  1.2\% &   6.1B & 0.95 \\
\midrule
Pretraining replay & $\sim$2.0T & 20.0\% & 100.7B & 0.05 \\
Long-context data  &  67.4B &  5.0\% &  25.2B & 0.37 \\
\bottomrule
\end{tabular}
\caption{Sources of the midtraining blend in Figure~\ref{fig:blend}. Replay is drawn from the base model's final-phase pretraining blend. Epochs are tokens seen divided by unique tokens.}
\label{tab:blend}
\end{table}

%% file: sections/4_results.tex
\section{Experiments and Results}
\label{sec:results}

\subsection{Evaluation protocol}
\label{sec:protocol}

The success of midtraining is ultimately measured by how well the model does after post-training. Our evaluation protocol is therefore holistic. It runs the full post-training pipeline and also tracks base-model post-trainability metrics and pretraining general-capability benchmarks.

\begin{itemize}
  \item \textbf{Short and full SFT.} We use an agentic-focused SFT blend (Appendix~\ref{app:sft}) at two budgets, 33B tokens (short SFT) and 100B tokens (full SFT), and compare each midtrained model with its starting checkpoint after identical SFT. Full SFT was close to saturation in our runs and supports the main claims, while short SFT gives a quicker signal.
  \item \textbf{Fixed evaluations.} We evaluate on SWE-bench Verified, Multilingual, and Pro, DeepSWE, Terminal-Bench (version 2.0, or 2.1 where noted), and a BrowseComp subset. Unless stated otherwise, SWE-bench runs in OpenHands and Terminal-Bench in Terminus-2, with pass@1 averaged over 3 and 8 rollouts per task (Appendix~\ref{app:eval} gives the setup and references). We also report pass@$k$, since a higher pass@$k$ has been shown to indicate more room for RL to improve \citep{yue2025rl}.
  \item \textbf{Base-model post-trainability metrics.} LoCoDiff \citep{locodiff}, RepoBench \citep{repobench}, and bits-per-byte on held-out code repositories and on agent traces (Appendix~\ref{app:bpb}) measure capabilities that midtraining targets, such as file-state tracking and cross-file modeling. For traces we score the decisive turns of DeepSWE trajectories \citep{beforesolve}, which correlated best with post-training results. LoCoDiff is averaged over 3 greedy runs on all 198 cases, and the other metrics are single runs. They improve with midtraining (Table~\ref{tab:base-metrics}), but did not track post-training results closely enough to guide recipe decisions.
  \item \textbf{Regression checks.} During midtraining we also track the pretraining general-capability benchmarks from \citet{nemotron3super} in four groups, general knowledge, math, code, and commonsense understanding (Appendix~\ref{app:eval}), to catch regressions (\S\ref{sec:replay}).
\end{itemize}

\subsection{Blend validation on \nano{}}
\label{sec:nano}

\begin{table}[t]
\centering
\small
\begin{tabular}{lccc}
\toprule
 & \textbf{SWE-V} & \textbf{SWE-Multi} & \textbf{TB~2.0} \\
 & pass@1 / pass@3 & pass@1 / pass@3 & pass@1 / pass@8 \\
\midrule
\rowcolor{gray!12}\multicolumn{4}{l}{\emph{Short SFT (33B tokens)}}\\
Base-8K                 & 57.1 / 69.2 & 48.6 / 65.0 & \textbf{22.1} / \textbf{40.5} \\
Midtrained, no traces   & 56.1 / 68.2 & 48.4 / 65.3 & 19.8 / \textbf{40.5} \\
Midtrained              & \textbf{57.9} / \textbf{71.4} & \textbf{56.8} / \textbf{70.0} & 20.1 / 39.3 \\
\rowcolor{gray!12}\multicolumn{4}{l}{\emph{Full SFT (100B tokens)}}\\
Base-8K                 & 59.7 / 69.8 & 55.9 / 72.7 & 22.9 / \textbf{44.9} \\
Midtrained, no traces   & 60.5 / 72.6 & 57.1 / 72.7 & 18.5 / 34.8 \\
Midtrained              & \textbf{63.5} / \textbf{73.2} & \textbf{59.6} / \textbf{74.0} & \textbf{25.1} / 43.8 \\
\bottomrule
\end{tabular}
\caption{Early experiment on the \nanoeightk{} checkpoint, midtrained for 100B tokens with and without agentic traces and compared with the base model after identical short or full SFT (\%).}

\label{tab:nano}
\vspace{-3mm}
\end{table}

Our first experiment midtrains the \nanoeightk{} base model for 100B tokens, once on the full blend and once without agentic traces, and compares both with the base model after identical SFT (Table~\ref{tab:nano}). Gains seen after short SFT carry over to full SFT, so short SFT can serve as a quick check. Midtraining improves SWE-bench Verified and Multilingual pass@1 by 0.8 and 8.2 points after short SFT, and both gains remain after full SFT (3.8 and 3.7 points). Terminal-Bench is noisier and improves only in full-SFT pass@1. Agentic traces give the majority of the gain after SFT. The no-traces model, trained on natural data alone, gains less than one point on SWE-bench Verified after full SFT, while the full blend improves on it by 3.0, 2.5, and 6.6 points on SWE-bench Verified, Multilingual, and Terminal-Bench. We still keep natural data as the large majority of the blend. Its benefits show up beyond SWE-bench pass@1, in file-state tracking and cross-file modeling (Tables~\ref{tab:base-metrics} and~\ref{tab:locodiff}) and robustness across agent harnesses (\S\ref{sec:harness}), and a blend of mostly traces would amount to a much longer SFT, which can hurt later RL trainability \citep{rejuvenation,quagmires,goodsft}.

\subsection{Scaling \nano{} to 500B tokens}
\label{sec:scaling}
Our main \nano{} experiment uses the long-context checkpoint \nanolc{} as the base model (\S\ref{sec:training}). The blend has about 380B unique tokens (\ifarxiv Table~\ref{tab:blend}\else Appendix~\ref{app:blend}\fi), enough to midtrain for 500B tokens while seeing each natural source about once. We apply identical full SFT to the base model and to midtrained checkpoints from 100B to 500B tokens (Table~\ref{tab:midtraining-scaling}).

Midtraining improves every benchmark over the base model. After 500B tokens, pass@1 rises by 3.8 points on SWE-bench Verified, 7.0 on SWE-bench Multilingual, 3.5 on SWE-bench Pro, and 1.0 on Terminal-Bench, and pass@$k$ rises on all four, by 2.3 to 4.7 points. Most of the SWE-bench Verified and Pro gain arrives by 100B tokens, SWE-bench Multilingual keeps improving up to 500B, and Terminal-Bench is noisier and peaks at 300B. Unlike Table~\ref{tab:nano}, these results start from \nanolc{}. After identical full SFT, its scores differ slightly from those of \nanoeightk{}, but stay within run-to-run variation. Midtraining also changes how the agent uses its context. Without midtraining, 4.1\% of SWE-bench Pro rollouts fail by filling the 256K context window, compared with 0.6\% after 500B tokens (Appendix~\ref{app:context}).

The base-model post-trainability metrics show the same trend (Table~\ref{tab:base-metrics}). All four improve over the base model and keep improving from 100B to 500B tokens, reflecting the capabilities that the natural SWE data and traces add. On LoCoDiff, for example, the base model rebuilds no files exactly, while the midtrained model rebuilds 21.5\% after 100B tokens and 26.8\% after 500B. Most of this gain comes from the diff data, which raises exact match from about 1\% to 30\% in a controlled 10B-token comparison in Table~\ref{tab:locodiff}.

\begin{table}[t]
\centering
\small
\begin{tabular}{ccccc}
\toprule
Midtraining & SWE-V & SWE-Multi & TB 2.0 & SWE-Pro \\
tokens (B) & pass@1 / pass@3 & pass@1 / pass@3 & pass@1 / pass@8 & pass@1 / pass@3 \\
\midrule
0 (base) & 60.7 / 72.2
    & 55.8 / 72.3
    & 22.6 / 43.8
    & 32.0 / 41.5 \\
100 & 64.1 / 74.0
    & 57.4 / 71.7
    & 22.5 / 44.9
    & 34.9 / 45.1 \\
200 & 64.3 / 73.4
    & 59.7 / 76.7
    & 21.4 / 39.3
    & 35.8 / 45.3 \\
300 & 63.5 / 73.6
    & 62.0 / 76.7
    & \textbf{24.9 / 48.3}
    & \textbf{36.3 / 46.4} \\
500 & \textbf{64.5 / 75.0}
    & \textbf{62.8 / 77.0}
    & 23.6 / 46.1
    & 35.5 / 45.4 \\
\bottomrule
\end{tabular}
\caption{Midtraining scaling on \nanolc{} after identical full SFT. After 500B tokens, midtraining improves both pass@1 and pass@$k$ on every benchmark, so it raises what the model can solve rather than only its reliability.}
\label{tab:midtraining-scaling}
\vspace{-3mm}
\end{table}

\begin{table}[t]
\centering
\small
\setlength{\tabcolsep}{4pt}
\begin{tabular}{lccccc}
\toprule
 & \multicolumn{2}{c}{LoCoDiff} & DeepSWE & Repository & RepoBench \\
\cmidrule(lr){2-3}
 & Exact (\%) $\uparrow$ & Distance $\downarrow$ & BPB $\downarrow$ & BPB $\downarrow$ & $\uparrow$ \\
\midrule
\nanolc{}          & 0.0 & 0.903 & 0.171 & 0.238 & 70.09 \\
Midtrained, 100B   & 21.5 & 0.228 & 0.144 & 0.228 & 71.22 \\
Midtrained, 500B   & \textbf{26.8} & \textbf{0.164} & \textbf{0.140} & \textbf{0.217} & \textbf{72.41} \\
\bottomrule
\end{tabular}
\caption{Base-model post-trainability metrics for \nanolc{} and its midtrained checkpoints from the scaling run. All metrics improve as midtraining scales.}
\label{tab:base-metrics}
\vspace{3mm}
\small
\begin{tabular}{lccc}
\toprule
Midtraining data & Tokens & Exact (\%) $\uparrow$ & Distance $\downarrow$ \\
\midrule
Blend without diff data & 10B & \phantom{0}1.3 & 0.507 \\
50\% diff data & 10B & \textbf{29.8} & \textbf{0.150} \\
Full blend (\S\ref{sec:scaling}) & 500B & 26.8 & 0.164 \\
\bottomrule
\end{tabular}
\caption{LoCoDiff on midtrained \nano{} base models. The two 10B-token runs start from \nanoeightk{} and differ only in whether half of the blend is diff data. Diff data drives most of the LoCoDiff gain.}
\label{tab:locodiff}
\vspace{-3mm}
\end{table}

\subsection{Scaling to \superm{}}
\label{sec:super}

We further apply the recipe to the larger \superm{} to test how it scales with model size. We midtrain for 65B and 280B tokens with the blend in Figure~\ref{fig:blend}. As in the scaling run, we start from the long-context checkpoint and compare each midtrained model with the base model after identical short and full SFT (Table~\ref{tab:super}). The clearest gain is on DeepSWE \citep{deepswe}, the hardest benchmark, whose long-horizon tasks are written from scratch. After full SFT, 280B tokens of midtraining raise it from 11.7 to 15.7, and after short SFT from 6.4 to 12.4. SWE-bench Verified also improves, from 72.9 to 74.3 after full SFT, and BrowseComp by 3 points after both short and full SFT. SWE-bench Multilingual and Pro do not improve. After full SFT they are 1.1 and 2.1 points below the base model, while after short SFT they are level with or slightly above it, so we do not read these differences as a consistent effect.

\begin{table}[t]
\centering
\small
\setlength{\tabcolsep}{4.5pt}
\begin{tabular}{lccccc}
\toprule
Midtraining & SWE-V & SWE-Multi & SWE-Pro & DeepSWE & BrowseComp \\
\midrule
\rowcolor{gray!12}\multicolumn{6}{l}{\emph{Short SFT (33B tokens)}}\\
0 (base) & 71.3 & 65.9 & 52.1 & \phantom{0}6.4 & 64.2 \\
65B      & 71.9 & 65.6 & 51.7 & 10.4 & 65.0 \\
280B     & \textbf{72.1} & \textbf{67.4} & \textbf{52.2} & \textbf{12.4} & \textbf{67.2} \\
\rowcolor{gray!12}\multicolumn{6}{l}{\emph{Full SFT (100B tokens)}}\\
0 (base) & 72.9 & \textbf{69.0} & \textbf{53.9} & 11.7 & 61.8 \\
65B      & 72.5 & 68.2 & 52.4 & 13.5 & 61.8 \\
280B     & \textbf{74.3} & 67.9 & 51.8 & \textbf{15.7} & \textbf{64.8} \\
\bottomrule
\end{tabular}
\caption{\superm{} after 65B and 280B midtraining tokens and identical short or full SFT (pass@1, \%). Midtraining gives a large gain on DeepSWE, the hardest benchmark, and also improves SWE-bench Verified and BrowseComp.}
\label{tab:super}
\end{table}

\subsection{Harness robustness}
\label{sec:harness}

Open models are used with many agent harnesses, not only the one their SFT data was built for, so harness robustness is a goal of our midtraining and the reason we draw traces from many scaffolds. Table~\ref{tab:harness} shows that midtraining improves pass@1 on every harness we tested, which makes both models more robust to the choice of harness, although large differences between harnesses remain. On \superm{} after short SFT, the midtrained model is better on every harness, by 8.6 points on average on SWE-bench Verified and 6.8 on Terminal-Bench~2.1. On \nanoeightk{} after full SFT, midtraining improves SWE-bench Verified on all six harnesses, by 11.6 points on average. Midtraining on the natural data alone already improves over the base model on four of the six harnesses, by up to 4.7 points, and adding the traces gives the largest gains. We did not train on traces from a single scaffold, so these results do not separate the effect of scaffold diversity from that of adding traces.

\begin{table}[t]
\centering
\small
\setlength{\tabcolsep}{3.6pt}
\begin{tabular}{l ccc ccc | cccc}
\toprule
 & \multicolumn{6}{c}{\superm{}, short SFT} & \multicolumn{4}{c}{\nanoeightk{}, full SFT} \\
\cmidrule(lr){2-7}\cmidrule(lr){8-11}
 & \multicolumn{3}{c}{SWE-bench Verified} & \multicolumn{3}{c}{Terminal-Bench 2.1} & \multicolumn{4}{c}{SWE-bench Verified} \\
\cmidrule(lr){2-4}\cmidrule(lr){5-7}\cmidrule(lr){8-11}
Harness & Base & Mid 65B & $\Delta$ & Base & Mid 65B & $\Delta$ & Base & \shortstack{Mid 100B\\[-1pt]{\scriptsize (no traces)}} & Mid 100B & $\Delta$ \\
\midrule
Pi          & 57.4 & 67.8 & \gain{10.4} & 30.3 & 35.2 & \gain{4.9}  & 45.7 & 48.8 & 59.9 & \gain{14.2} \\
OpenCode    & 54.4 & 65.8 & \gain{11.4} & 35.2 & 38.2 & \gain{3.0}  & 52.0 & 50.1 & 63.1 & \gain{11.1} \\
Hermes      & 59.8 & 63.2 & \gain{3.4}  & 44.2 & 53.9 & \gain{9.7}  & 43.3 & 43.3 & 50.0 & \gain{6.7} \\
Codex       & 48.4 & 61.8 & \gain{13.4} & 27.0 & 39.7 & \gain{12.7} & 42.1 & 45.9 & 61.1 & \gain{19.0} \\
Claude Code & 53.8 & 58.0 & \gain{4.2}  & 39.7 & 43.5 & \gain{3.8}  & 46.0 & 50.7 & 60.9 & \gain{14.9} \\
OpenHands   & --   & --   & --          & --   & --   & --          & 59.7 & 60.5 & 63.5 & \gain{3.8} \\
\midrule
Mean        & 54.8 & \textbf{63.3} & \gain{8.6} & 35.3 & \textbf{42.1} & \gain{6.8} & 48.1 & 49.9 & \textbf{59.8} & \gain{11.6} \\
\bottomrule
\end{tabular}
\caption{Midtraining improves pass@1 on every agent harness we evaluated, for both models, with the largest gain on the harness where the base model is weakest (\%).}
\label{tab:harness}
\vspace{-3mm}
\end{table}

%% file: sections/5_analysis.tex
\section{Analysis and Ablations}
\label{sec:analysis}

\subsection{Midtraining as a long-context stage}
\label{sec:lc}

Because midtraining trains at 256K, it also extends the model's context. Table~\ref{tab:nano_ruler} compares the usual order, where \nanoeightk{} goes straight to the long-context stage, with a 100B-token \nano{} midtraining run at 256K before a long-context stage. Midtraining alone brings \nanoeightk{} within about five points of the long-context base model on RULER \citep{ruler} up to 128K, but trails by 14 points at 256K. After the long-context stage, the midtrained model beats the long-context base model at every length from 64K to 256K and stays within two points at shorter lengths. Midtraining thus does part of the work of the long-context stage, and this carries over to the final model. Appendix~\ref{app:lc} gives more results for \superm{}.

\begin{table}[t]
\centering
\small
\begin{tabular}{ccccccc}
\toprule
 & 8K & 16K & 32K & 64K & 128K & 256K \\
\midrule
8K $\rightarrow$ long-context @1M                   & \textbf{94.1} & \textbf{91.9} & \textbf{90.5} & 85.9 & 81.3 & 76.6 \\
8K $\rightarrow$ midtraining @256K                   & 92.3 & 91.4 & 87.8 & 81.9 & 76.1 & 62.2 \\
8K $\rightarrow$ midtraining @256K $\rightarrow$ long-context @1M & 92.1 & 91.0 & 89.8 & \textbf{86.6} & \textbf{83.8} & \textbf{79.7} \\
\bottomrule
\end{tabular}
\caption{\nano{} RULER scores up to 256K. Midtraining alone brings \nanoeightk{} close to \nanolc{} up to 128K, acting like a long-context stage.}
\label{tab:nano_ruler}
\vspace{-3mm}
\end{table}

\subsection{Masking tool responses in traces}
\label{sec:masking}

Agent traces interleave the model's own actions with tool responses, the outputs the environment returns after each tool call. Some recent work also trains on tool responses, so that the model learns to predict how the environment reacts, a form of world modeling \citep{cwm,davincidev}.

\ifarxiv
\begin{table}[t]
\centering
\small
\begin{tabular}{lcc}
\toprule
SWE-V & pass@1 & pass@3 \\
\midrule
Unmasked & 27.5 & 41.6 \\
Masked & \textbf{37.8} & \textbf{53.0} \\
\bottomrule
\end{tabular}
\caption{Tool-response masking. \nanoeightk{} midtrained for 100B tokens with and without tool responses in the loss, followed by identical SFT with an earlier version of our SFT recipe (SWE-bench Verified).}
\label{tab:tool-response-masking}
\vspace{-3mm}
\end{table}
\else
\begin{wraptable}{r}{0.42\textwidth}
\vspace{-3mm}
\centering
\small
\begin{tabular}{lcc}
\toprule
SWE-V & pass@1 & pass@3 \\
\midrule
Unmasked & 27.5 & 41.6 \\
Masked & \textbf{37.8} & \textbf{53.0} \\
\bottomrule
\end{tabular}
\caption{Tool-response masking.}
\label{tab:tool-response-masking}
\vspace{-2mm}
\end{wraptable}
\fi
We tested this with the setup of \S\ref{sec:nano}, midtraining \nanoeightk{} for 100B tokens with and without tool responses in the loss, followed by identical SFT with an earlier version of our SFT recipe. In our experiments, training on tool responses did not work out of the box and lowered SWE-bench Verified pass@1 by about 10 points (Table~\ref{tab:tool-response-masking}), so we train only on assistant turns, as in SFT. Tool responses are often long and noisy, such as file dumps, test logs, and error traces, and predicting them may take capacity away from learning the agent's own actions.

\subsection{Pretraining replay and learning rate}
\label{sec:replay}
\label{sec:lr}

Replay keeps the model close to its pretraining distribution, so that midtraining shifts the data distribution gradually rather than abruptly. We started from 20\% replay as a reasonable default, with the goal of no regression on general-capability benchmarks. Table~\ref{tab:super-base} shows that \superm{} does not regress through 65B tokens, with every benchmark group within 1.2 points of the starting checkpoint. We also tried 40\% replay, which gave no clear benefit over 20\%, so we kept 20\%.

\begin{table}[t]
\centering
\footnotesize
\begin{minipage}[t]{0.55\textwidth}
\centering
\setlength{\tabcolsep}{4pt}
\begin{tabular}[t]{lcccc}
\toprule
Midtraining & Knowledge & Math & Code & Commonsense \\
\midrule
0 (base) & 79.7 & 86.4 & \textbf{80.8} & \textbf{87.5} \\
25B      & \textbf{79.9} & 85.9 & 80.0 & 87.1 \\
65B      & 79.8 & \textbf{87.0} & 79.6 & 86.7 \\
\bottomrule
\end{tabular}
\end{minipage}\hfill
\begin{minipage}[t]{0.42\textwidth}
\centering
\setlength{\tabcolsep}{4pt}
\begin{tabular}[t]{lccc}
\toprule
LR & MMLU & MMLU-Pro & MATH-500 \\
\midrule
Base & \textbf{79.4} & 64.3 & \textbf{79.7} \\
1e-5 & 78.5 & \textbf{66.3} & 78.8 \\
3e-5 & 76.7 & 61.5 & 74.3 \\
5e-5 & 76.8 & 63.3 & 61.7 \\
\bottomrule
\end{tabular}
\end{minipage}
\\[2pt]
\small
\begin{minipage}[t]{0.55\textwidth}
\caption{\superm{} pretraining benchmarks before and after midtraining with 20\% replay.}
\label{tab:super-base}
\end{minipage}\hfill
\begin{minipage}[t]{0.42\textwidth}
\caption{\nanoeightk{} after 100B midtraining tokens at three learning rates.}
\label{tab:lr-ablation}
\end{minipage}
\vspace{-3mm}
\end{table}

The learning rate also affects general capabilities. Pretraining ends with a small, decayed learning rate, and midtraining needs a higher one to learn the new data. In our early runs, midtraining at 1e-5 gave smaller gains after SFT than higher rates did. Raising the rate too far, however, hurts general capabilities even with replay. Table~\ref{tab:lr-ablation} compares a constant 1e-5 with warming up to 3e-5 or 5e-5. Both higher rates lower general capabilities, most clearly MATH-500, which falls from 79.7 to 61.7 at 5e-5. We therefore recommend a moderate increase and checking general-capability benchmarks while tuning it.

%% file: sections/6_discussion.tex
\section{Conclusion}
\label{sec:discussion}

We presented an open midtraining recipe for agentic coding, with full details of how each data source is built, filtered, and weighted and how the models are trained. After identical SFT, the recipe gives clear gains as midtraining scales to 500B tokens on \nano{}, and it carries over to the larger \superm{}, where it raises DeepSWE, the hardest benchmark we evaluate, from 11.7 to 15.7 and also improves SWE-bench Verified and BrowseComp. On both models, midtraining improves results on every agent harness we tested. Our experiments show that agentic traces are essential, since natural software-engineering data alone gives much smaller gains after SFT, and that a short SFT run gives a more direct post-training signal than base-model post-trainability metrics, which remain useful to track. Midtraining at 256K also works as an intermediate long-context stage, and we explain our other modeling choices, such as tool-response masking and pretraining replay.

\paragraph{Limitations and future work.}
Agentic benchmarks are noisy, and each comparison is a single run, so differences of one or two points, especially on Terminal-Bench, may come from run-to-run variation. We did not compare against spending the same tokens on longer pretraining or SFT, since midtraining adds a separate stage that complements SFT rather than replacing it. We tested the recipe only on Nemotron models, not on other model families, and mainly on agentic coding. Our SFT evaluations use an agentic-focused SFT blend rather than a full general-purpose post-training blend. Our RL results are also limited, since clear RL gains on top of a strong SFT checkpoint were difficult to get at the \nano{} scale and RL was too compute intensive at the \superm{} scale, so we report only a short, single-seed run on \nanoeightk{} (Appendix~\ref{sec:rl}). For future work, since traces drove most of the gains, we plan to scale trace data with more diversity, and we have already expanded our trace corpus to about 90B tokens. We also plan to extend the agentic data to more capabilities and to move more post-training data, such as reasoning and chat data, into midtraining.

%% file: sections/7_appendix.tex
\section{Midtraining data}
\label{app:blend}
\ifarxiv
This appendix documents every source in the blend, including how it is collected, filtered, and weighted, and Appendix~\ref{app:training} lists the full training settings. We plan to release data-processing scripts and training-set metadata in a future version.
\fi

\ifarxiv\else
\input{sections/blend_table}
\fi

\begin{figure}[t]
\centering
\includegraphics[width=0.9\textwidth]{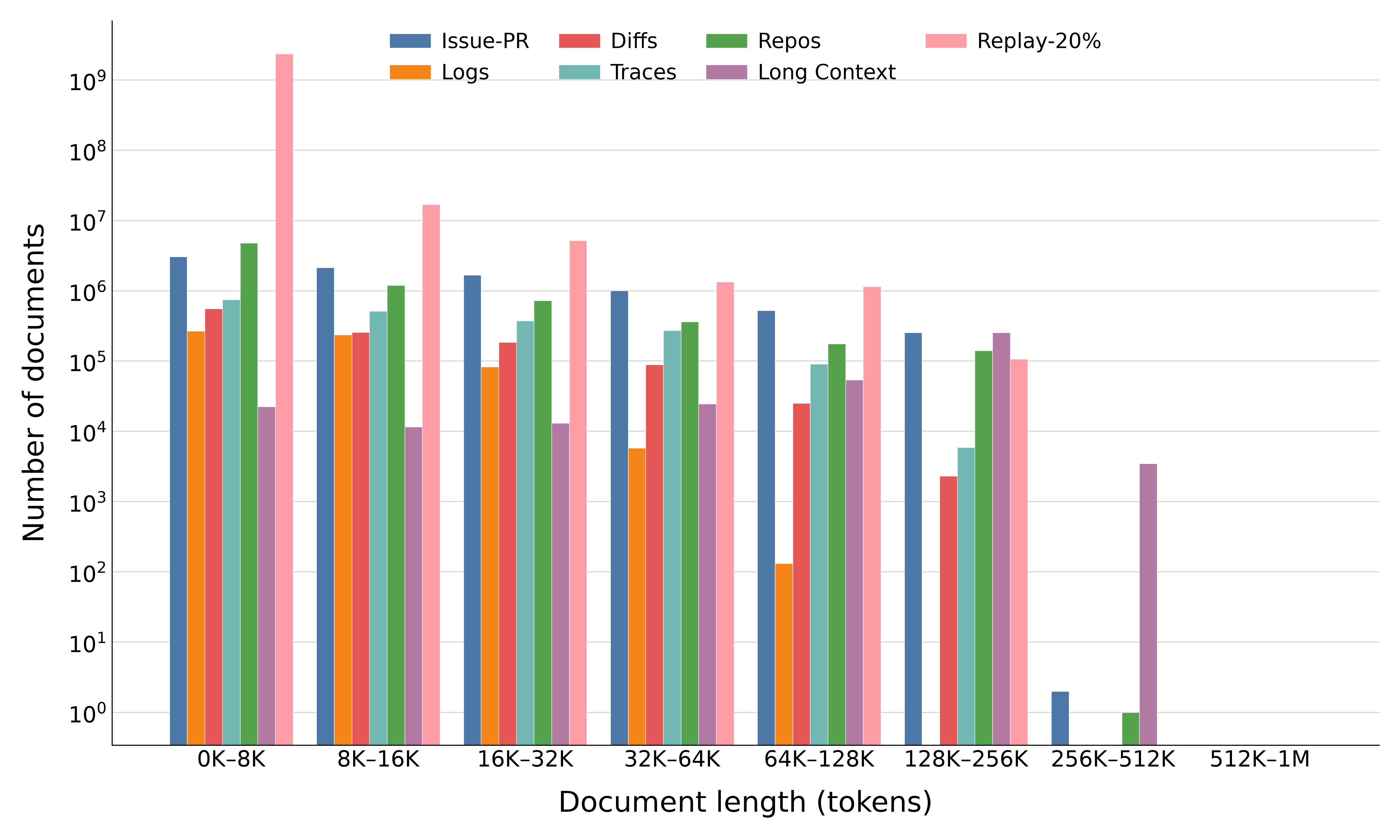}
\caption{Document lengths in tokens for each source in the midtraining blend (log scale).}
\label{fig:doc-lengths}
\end{figure}

\subsection{Issue--PR}
\label{app:issuepr}
For PRs linked to an issue, an LLM judge checks whether the issue and the PR match. The judge reads the full issue and PR text, the final patch, and the paths of the changed files, and labels each pair as clean or assigns one of seven failure modes, namely a revert, an incidental issue link, a wrong issue reference, a partial fix, an issue with no clear action, an underspecified issue, or a PR that does much more than the issue asks. We keep clean pairs and remove pairs that cannot be fixed. We repair the others by linking the PR to the correct issue, narrowing its scope, keeping it as a PR-only example, or splitting it from unrelated changes. For PRs without a linked issue, we remove those that change only README files or three lines or fewer.

\subsection{Repositories}
\label{app:repos}
The repository data starts from the code repositories in the Nemotron 3 pretraining corpus \citep{nemotron3}, a deduplicated collection of public repositories with permissive licenses, and covers 14 languages (Table~\ref{tab:repo-langs}).

\paragraph{Connectedness.} For each language, we parse the references that link one file to another, such as imports in Python, includes in C and C++, and module paths in Go. A repository's connected fraction is the share of its code files that import another file in the repository, and we keep repositories where this share is at least 20\%. For Ruby we lower the threshold to 10\%, because Rails autoloading hides most references from static parsing. We exclude languages with little cross-file structure that parsing can recover, such as Shell, R, and Swift, as well as markup and configuration languages.

\paragraph{Serialization.} The code files follow the README in depth-first order over the import graph, so the files of each connected component stay together and each file sits near the files it imports. Each file starts with a header that gives the repository name and file path, and the header is masked from the loss.

\paragraph{Filtering.} We deduplicate files by path within each repository and drop Python~2 code, removing the whole repository when most of it is Python~2. We drop repositories that contain server-side secrets, such as cloud or API keys, and redact client-side keys, which are common in mobile and web projects. We also strip license and auto-generated headers, and drop repositories above 2M tokens or those that lose most of their files in deduplication. To avoid contamination, we remove SWE-bench repositories and their forks, as well as files that contain HumanEval, MBPP, or LiveCodeBench \citep{livecodebench} solutions.

\paragraph{Blend weights.} We start from each language's natural token share, upweight Python, and downweight Dart, Scala, and Lua (Table~\ref{tab:repo-langs}).

\begin{table}[h]
\centering
\small
\begin{tabular}{lrrrr}
\toprule
Language & Records & Tokens (B) & Natural share (\%) & Blend weight (\%) \\
\midrule
Python     & 1,798,331 & 24.51 & 23.5 & 31.0 \\
Java       &   945,513 & 18.71 & 18.0 & 17.0 \\
TypeScript & 1,273,349 & 13.20 & 12.7 & 12.0 \\
C\#        &   460,597 & 11.66 & 11.2 & 10.0 \\
JavaScript & 1,289,843 & 10.01 &  9.6 &  8.6 \\
C++        &   400,540 &  8.27 &  7.9 &  7.6 \\
PHP        &   329,934 &  5.30 &  5.1 &  4.7 \\
Dart       &   386,648 &  3.24 &  3.1 &  1.4 \\
C          &   106,799 &  2.81 &  2.7 &  2.7 \\
Rust       &   115,314 &  2.22 &  2.1 &  1.9 \\
Go         &   119,005 &  1.68 &  1.6 &  1.5 \\
Ruby       &    89,846 &  1.00 &  1.0 &  0.9 \\
Scala      &    39,618 &  0.99 &  1.0 &  0.4 \\
Lua        &    21,770 &  0.51 &  0.5 &  0.3 \\
\midrule
Total      & 7,377,107 & 104.12 & 100.0 & 100.0 \\
\bottomrule
\end{tabular}
\caption{Repository data by language. Tokens are measured with the model tokenizer.}
\label{tab:repo-langs}
\end{table}

\subsection{Diff data}
\label{app:diff}
Multi-file edits come from merged pull requests in the Issue--PR corpus, with the full patch and every changed file before the change. Code-evolution documents come from the full git histories of 6,550 permissively licensed repositories from the same corpus, chosen to match the language mix of SWE-bench Multilingual. We keep only patches that apply exactly and histories that replay byte for byte. In multi-file edits, the loss covers the original files, the diffs, and the changed lines of the new files with two lines of context. In code-evolution documents, it covers the commit messages, diffs, and final state, and license blocks are masked in both. We remove all repositories used in SWE-bench Verified, SWE-bench Pro, Multi-SWE-bench \citep{multiswebench}, Terminal-Bench, and LoCoDiff, deduplicate examples, and drop GPL-licensed documents, secrets, lockfiles, and generated files. Histories have a median of 3 commits and up to 59, and the median document is about 6K tokens (Table~\ref{tab:diff-stats}).

\begin{table}[h]
\centering
\small
\begin{tabular}{lrrrr}
\toprule
Type & Examples & Document tokens & Trained tokens & Trained share \\
\midrule
Multi-file edits & 733,705 & 13.3B & 6.8B & 51\% \\
Code evolution   & 377,618 &  3.5B & 3.2B & 91\% \\
\midrule
Total            & 1,111,323 & 16.8B & 10.0B & 60\% \\
\bottomrule
\end{tabular}
\caption{Diff data used in our experiments. Trained tokens exclude masked context. Counts use the tokenizer of our data pipeline. With the model tokenizer, the total is 16.0B (Table~\ref{tab:blend}).}
\label{tab:diff-stats}
\end{table}

\subsection{Build and CI logs}
\label{app:logs}
Raw logs come from the full LogHub archives (BGL, Thunderbird, HPC, HDFS, Spark, and Windows). We cut them into non-overlapping windows of 120 context lines followed by 16 target lines. We deduplicate HDFS logs after masking block identifiers, hosts, and timestamps. No single archive is more than 28\% of this source. Supercomputer logs are at least 30\%, and CI/CD logs from GHALogs run logs and workflow-file histories are at least 12\%. We check every source against a license registry and exclude sources with unknown licenses. We also build 632 diagnosis examples. Each centers a window of HPC logs on a warning, error, or critical event, and a frontier model writes the diagnosis from the window and from findings that simple rules extract. We redact secrets and mask timestamps, hostnames, hashes, and repeated boilerplate from the loss.

\subsection{Agentic traces}
\label{app:traces}
The blend in Table~\ref{tab:blend} uses 39.4B tokens of traces. The statistics below come from an early filtered snapshot of 859,176 trajectories. Scaffold labels name the agent interface, not the task domain. For example, a software-engineering task can use shell commands, file-editing tools, or both. We downsample general tool-use data, mainly \texttt{taubench\_react}, to give more weight to software engineering. Tables~\ref{tab:trace-scaffolds}, \ref{tab:trace-tools}, and~\ref{tab:trace-turns} give the scaffolds, the most frequent tools, and the number of assistant turns per trajectory. In the snapshot, filtering (\S\ref{sec:recipe}) removed 8,935 trajectories, about 1\%, mostly unfinished runs (6,960) or repeated loops that ended in failure (1,959).

\begin{table}[t]
\centering
\small
\begin{tabular}{lrr}
\toprule
Scaffold label & Trajectories & Share (\%) \\
\midrule
\texttt{codex}             & 166,624 & 19.39 \\
\texttt{openhands}         & 120,086 & 13.98 \\
\texttt{generic\_code\_cli} & 107,291 & 12.49 \\
\texttt{gemini\_cli}       & 104,717 & 12.19 \\
\texttt{terminus}          &  99,915 & 11.63 \\
\texttt{taubench\_react}   &  50,399 &  5.87 \\
\texttt{swe\_agent}        &  46,162 &  5.37 \\
\texttt{opencode}          &  44,920 &  5.23 \\
\texttt{mini\_swe\_agent}  &  35,260 &  4.10 \\
\texttt{search\_browse}    &  26,168 &  3.05 \\
\texttt{unknown}           &  25,866 &  3.01 \\
\texttt{chat}              &  20,426 &  2.38 \\
\texttt{claude\_code}      &  11,342 &  1.32 \\
\midrule
Total                     & 859,176 & 100.00 \\
\bottomrule
\end{tabular}
\caption{Trajectories by scaffold label in the early snapshot. Shares are of trajectories, not tokens.}
\label{tab:trace-scaffolds}
\end{table}

\begin{table}[t]
\centering
\small
\begin{tabular}{lrr}
\toprule
Tool name & Calls & Share (\%) \\
\midrule
\texttt{bash}                 & 3,659,199 & 19.6 \\
\texttt{execute\_bash}        & 3,633,328 & 19.4 \\
\texttt{str\_replace\_editor} & 2,913,745 & 15.6 \\
\texttt{keystrokes}           & 2,029,095 & 10.9 \\
\texttt{exec\_command}        &   736,655 &  3.9 \\
\texttt{write}                &   524,113 &  2.8 \\
\texttt{browser.search}       &   317,169 &  1.7 \\
\texttt{shell}                &   280,803 &  1.5 \\
\texttt{read}                 &   261,285 &  1.4 \\
\texttt{todowrite}            &   260,147 &  1.4 \\
\bottomrule
\end{tabular}
\caption{The ten most frequent tool names in the early snapshot, as each scaffold records them, so similar tools such as \texttt{bash} and \texttt{execute\_bash} are listed separately. Shares are of all tool calls.}
\label{tab:trace-tools}
\end{table}

\begin{table}[t]
\centering
\small
\begin{tabular}{lrr}
\toprule
Assistant turns & Trajectories & Share (\%) \\
\midrule
1         & 146,334 & 17.03 \\
2         &  56,929 &  6.63 \\
3         &  47,362 &  5.51 \\
4--5      &  77,700 &  9.04 \\
6--10     & 138,036 & 16.07 \\
11--20    & 111,777 & 13.01 \\
21--50    & 161,295 & 18.77 \\
51--100   & 105,711 & 12.30 \\
101--200  &  12,833 &  1.49 \\
201--300  &   1,195 &  0.14 \\
301--400  &       4 & $<0.01$ \\
401+      &       0 &  0.00 \\
\midrule
Total     & 859,176 & 100.00 \\
\bottomrule
\end{tabular}
\caption{Trajectory length in assistant turns, in the early snapshot.}
\label{tab:trace-turns}
\end{table}

\section{Training setup}
\label{app:training}
Table~\ref{tab:hparams} lists the settings for midtraining and SFT. All runs use Megatron-LM \citep{shoeybi2019megatron} on NVIDIA GB300 GPUs, with inter-document masking so that documents packed into the same sequence do not attend to each other. Both models keep their two multi-token-prediction (MTP) layers during midtraining.

\begin{table}[h]
\centering
\small
\setlength{\tabcolsep}{3.5pt}
\begin{tabular}{llll}
\toprule
 & \nano{} midtraining & \superm{} midtraining & SFT (both models) \\
\midrule
Parameters (total / active) & 30B / 3B & 120B / 12B & -- \\
Sequence length (packed) & 256K & 256K & 256K \\
Global batch size (sequences) & 64 & 64 & 64 \\
Peak learning rate & $5\times10^{-5}$ & $1\times10^{-5}$ & $2\times10^{-5}$ \\
Warmup (sequences) & 32K & 30K & -- \\
Schedule after warmup & constant & constant & cosine to $5\times10^{-6}$ \\
Optimizer & \multicolumn{3}{l}{Adam, $\beta=(0.9, 0.95)$, weight decay 0.1, gradient clip 1.0} \\
Precision & \multicolumn{3}{l}{BF16} \\
Parallelism (TP / CP / EP) & 2 / 4 / 8 & 2 / 8 / 64 & -- \\
MTP layers (loss weight) & 2 (0.1) & 2 (0.3) & -- \\
Tokens & up to 500B & 65B and 280B & 33B or 100B \\
\bottomrule
\end{tabular}
\caption{Training settings. TP, CP, and EP are tensor, context, and expert parallelism, and MTP is multi-token prediction.}
\label{tab:hparams}
\end{table}

\subsection{SFT data}
\label{app:sft}
Both models use the same SFT data and settings. Our SFT evaluations use an agentic-focused SFT blend of about 430K trajectories from several scaffolds, most of them generated by GLM-5.2. They cover software engineering, terminal use, search, conversational tool use, and knowledge work.

\section{Evaluation}
\subsection{Agentic benchmarks}
\label{app:eval}
For all agentic benchmarks, we sample with temperature 1.0 and top-p 0.95. SWE-bench \citep{swebench} Verified \citep{swebenchverified} (500 tasks), SWE-bench Multilingual \citep{swesmith} (300), and SWE-bench Pro \citep{swebenchpro} (731) use the OpenHands agent \citep{openhands,openhandssdk} (SDK version 1.17.0) with 3 rollouts per task. Each rollout has at most 200 turns, or 250 for SWE-bench Pro. Terminal-Bench 2.0 \citep{terminalbench} (89 tasks) uses the Terminus-2 agent \citep{terminalbench} with a one-hour limit per task and 8 rollouts per task. The harness sweep and the RL evaluation use Terminal-Bench~2.1, which has the same 89 tasks with 28 of them fixed, and we compare scores only within a version. DeepSWE \citep{deepswe} pass@1 is averaged over 4 runs. BrowseComp \citep{browsecomp} is evaluated on a subset of the benchmark in a custom agent harness, with Exa as the search provider. For \superm{}, run-to-run variation is about 1 point on SWE-bench Verified and Pro and about 2 points on DeepSWE. We report pass@1 as the mean success rate over rollouts and compute pass@$k$ from the same rollouts.

\paragraph{General-capability benchmarks.} The regression checks use the base-model evaluation suite from \citet{nemotron3super}. General knowledge averages MMLU \citep{mmlu} (5-shot) and MMLU-Pro \citep{mmlupro} (5-shot, chain of thought). Math averages GSM8K \citep{gsm8k} (8-shot) and MATH \citep{math} (4-shot). Code averages HumanEval \citep{humaneval} (0-shot) and MBPP-Sanitized \citep{mbpp} (3-shot). Commonsense understanding averages ARC-Challenge \citep{arc}, HellaSwag \citep{hellaswag}, OpenBookQA \citep{openbookqa}, PIQA \citep{piqa}, and WinoGrande \citep{winogrande}.

\subsection{Base-model post-trainability metrics}
\label{app:bpb}
We track four base-model post-trainability metrics during midtraining.
\begin{itemize}
  \item \textbf{LoCoDiff} \citep{locodiff} gives the model the full git history of a file and asks it to write the file's final state. We use all 198 cases, zero-shot with greedy decoding and a 16K-token generation budget, which covers the longest reference file (11.7K tokens). Exact match is the share of files reproduced byte for byte, and normalized edit distance is capped at 1.2 per case. Greedy decoding still varies slightly across runs because inference is not bit-exact, so we average both over 3 runs.
  \item \textbf{RepoBench} \citep{repobench} measures next-line code completion that requires context from other files in the repository. We report edit similarity in the cross-file-first setting, with greedy decoding.
  \item \textbf{Repository bits-per-byte} follows Composer~2 \citep{composer2}. We concatenate the files of held-out repositories and compute bits-per-byte over long windows. Our version puts the README first, orders files by dependency, keeps the file tree across windows, and does not score file headers.
  \item \textbf{Trace bits-per-byte} is measured on the assistant turns of held-out agent traces, following the decisive-action bits-per-byte of \citet{beforesolve}. The version in Table~\ref{tab:base-metrics} scores only the decisive turns of DeepSWE traces, the turns that most affect the outcome, and masks tool-call formatting, which otherwise dominates the score.
\end{itemize}

\section{Persistence of gains through reinforcement learning}
\label{sec:rl}

As a preliminary check on whether midtraining gains persist under RL, we run an identical short stage of RL with verifiable rewards (RLVR), 20 GRPO steps \citep{shao2024deepseekmath} over 14 verifiable environments (Appendix~\ref{app:rl}), from \nanoeightk{} and its midtrained checkpoint after full SFT. Twenty steps is too short to measure an RL gain, so we only ask whether the midtrained model's advantage survives the start of RL.

\subsection{Results}

\begin{table}[ht]
\centering
\small
\begin{tabular}{lcccc}
\toprule
SWE-V (pass@1 / pass@3) & SFT & RL step 10 & Step 15 & Step 20 \\
\midrule
Base-8K    & 61.2 / 71.8 & 60.4 / 70.8 & 61.4 / 72.2 & 60.4 / 71.7 \\
Midtrained-100B & \textbf{63.2 / 74.1} & \textbf{63.0 / 74.0} & \textbf{63.5 / 73.7} & \textbf{63.0 / 73.0} \\
\midrule
TB~2.1 (pass@1 / pass@8) & SFT & RL step 10 & Step 15 & Step 20 \\
\midrule
Base-8K    & 23.0 / \textbf{46.1} & \textbf{24.7} / 44.9 & 22.9 / 41.6 & 20.8 / 39.3 \\
Midtrained-100B & \textbf{24.2} / 44.9 & 23.5 / 44.9 & \textbf{24.4 / 44.9} & \textbf{23.2 / 46.1} \\
\bottomrule
\end{tabular}
\caption{RLVR from \nanoeightk{} and its midtrained checkpoint after full SFT. SFT-start scores are re-evaluated with the RL checkpoints. The midtrained model leads on SWE-bench, and its Terminal-Bench pass@8 holds.}
\label{tab:rl}
\end{table}

The midtrained model starts about two points ahead on SWE-bench Verified pass@1 and keeps that lead at every RL step, and both models stay close to their SFT scores (Table~\ref{tab:rl}). On Terminal-Bench~2.1, the base model's pass@8 falls from 46.1 to 39.3 over 20 steps while the midtrained model's holds at 45 to 46, although with 89 tasks this drop amounts to only six tasks. Training-time signals, including mean reward, generation length, and format violations, also favor the midtrained model (Appendix~\ref{app:rl}).

With a single seed and 20 steps, these results do not show that midtraining increases what RL can add, only that the midtrained model's advantage after SFT is not lost when RL begins.

\subsection{Setup and training signals}
\label{app:rl}
The RLVR prompt blend is drawn from 14 verifiable environments (tool use,
code generation, SWE and search pivots, instruction following, reasoning,
and formatting). We reward-profile a seed blend against both checkpoints
and keep only prompts with mixed rewards, meaning that for each checkpoint
some but not all of its sampled rollouts succeed, so every prompt is
neither trivial nor unsolvable for either model. Training is GRPO with a
leave-one-out baseline and no reference-policy KL penalty, 256 prompts
$\times$ 16 generations per step, a 32K-token output cap within a 131K
context, a learning rate of $10^{-6}$ with 10 warmup steps, and 20
optimizer updates. The two runs differ only in their starting checkpoint;
they use the same blend, seed, and prompt order. We evaluate SWE-bench Verified with 5 rollouts per task and Terminal-Bench~2.1 with 8, for both the SFT starting points and the RL checkpoints.

Training-time metrics, averaged over matched steps, favor the midtrained
model. It earns 4.3\% higher train reward (0.308 vs.\ 0.295) with 12\%
shorter generations (4.6K vs.\ 5.3K mean tokens), and it produces format
violations (a stray end-of-sequence token or malformed thinking tags, which
zero the reward) about ten times less often (0.03\% vs.\ 0.34\% of
trajectories). Its truncation rate is also lower (0.056\% vs.\ 0.073\%).
Together these make RL from the midtrained checkpoint about 15\% faster
per trajectory on the same hardware.

\section{Additional results}
\subsection{Context-limit failures}
\label{app:context}
Rollouts that fill the 256K context window count as failures. Under the same harness and settings, they are much more common without midtraining (Table~\ref{tab:context-fails}), especially on SWE-bench Pro. On SWE-bench Verified, rollouts also hit the 200-turn cap 11 times without midtraining and once after 500B tokens. Midtraining appears to make the agent more economical with its context and turns, not only more accurate.

\begin{table}[h]
\centering
\small
\begin{tabular}{lrcc}
\toprule
Benchmark & Rollouts & \nanolc{} & Midtrained, 500B \\
\midrule
SWE-bench Verified     & 1{,}500 & 40 & 26 \\
SWE-bench Multilingual &   900 & 25 & 14 \\
SWE-bench Pro          & 2{,}193 & 91 & 13 \\
Terminal-Bench         &   712 &  8 &  1 \\
\bottomrule
\end{tabular}
\caption{Rollouts that fail by filling the 256K context window, after identical full SFT.}
\label{tab:context-fails}
\end{table}

\subsection{Long-context results at 1M}
\label{app:lc}
To keep evaluation simple, our \superm{} runs start midtraining from the 1M long-context checkpoint (\S\ref{sec:training}). Midtraining then trains at 256K, so RULER at 1M naturally drops, from 79.9 to 33.0 after 25B tokens. A short long-context stage of about 9B tokens, on long-context data only, brings it back to 81.0, above the starting checkpoint, and the 65B \superm{} checkpoint evaluated in \S\ref{sec:super} includes this short stage, while the 280B checkpoint goes straight to SFT. This drop is a side effect of our evaluation setup. In a mature pipeline, midtraining would come before the final long-context stage, as in \S\ref{sec:lc}.